\documentclass[sigconf]{acmart}
\AtBeginDocument{%
  }

\copyrightyear{2026}
\acmYear{2026}
\setcopyright{cc}
\setcctype{by}
\acmConference[MM '26]{Proceedings of the 34th ACM International Conference on Multimedia}{November 10--14, 2026}{Rio de Janeiro, Brazil}
\acmBooktitle{Proceedings of the 34th ACM International Conference on Multimedia (MM '26), November 10--14, 2026, Rio de Janeiro, Brazil}
\acmDOI{10.1145/3767308.3835215}
\acmISBN{979-8-4007-2213-4/2026/11}

\begin{document}

%%
%% The "title" command has an optional parameter,
%% allowing the author to define a "short title" to be used in page headers.
\title{DyFrDet: Towards Accurate Small Object Detection via Dynamic Frequency Suppression with Label Disambiguation}

%%
%% The "author" command and its associated commands are used to define
%% the authors and their affiliations.
%% Of note is the shared affiliation of the first two authors, and the
%% "authornote" and "authornotemark" commands
%% used to denote shared contribution to the research.
\author{Zihan Yang}
\authornote{Both authors contributed equally to this research.}
\email{zihanyang@buaa.edu.cn}
\orcid{0009-0003-8818-5360}
\affiliation{
  \institution{Hangzhou International Innovation Institute of Beihang University}
  \city{Hangzhou}
  \country{China}
}

\author{Yang Guo}
\authornotemark[1]
\email{guoyang4409@gmail.com}
\orcid{0009-0000-0455-3217}
\affiliation{
  \institution{Beijing University of Posts and Telecommunications}
  \city{Beijing}
  \country{China}
}

\author{Hongxing Zhang}
\email{hongxingzhang@bupt.edu.cn}
\orcid{0000-0001-6637-5655}
\affiliation{
  \institution{Beijing University of Posts and Telecommunications}
  \city{Beijing}
  \country{China}
}

\author{Dan Lu}
\authornote{Corresponding authors: Dan Lu and Siyuan Yao.}
\email{danlu@buaa.edu.cn}
\orcid{0000-0002-9274-9398}
\affiliation{
  \institution{Hangzhou International Innovation Institute of Beihang University}
  \city{Hangzhou}
  \country{China}
}

\author{Siyuan Yao}
\authornotemark[2]
\email{yaosiyuan04@gmail.com}
\orcid{0000-0002-8479-5124}
\affiliation{
  \institution{Shenzhen Campus of Sun Yat-sen University}
  \city{Shenzhen}
  \country{China}
}

%%
%% By default, the full list of authors will be used in the page
%% headers. Often, this list is too long, and will overlap
%% other information printed in the page headers. This command allows
%% the author to define a more concise list
%% of authors' names for this purpose.
\renewcommand{\shortauthors}{Zihan Yang, Yang Guo, Hongxing Zhang, Dan Lu, \& Siyuan Yao.}

%%
%% The abstract is a short summary of the work to be presented in the
%% article.
\begin{abstract}
 Despite the remarkable progress over the past decades, accurately identifying small objects remains challenging because of their insufficient visual cues. Previous works typically attempt to construct discriminative representation of the small objects. However, the wide range frequency domain noises and label ambiguities have been greatly overlooked, which significantly hinders the accurate localization. To address these issues, we propose a novel small object detection (SOD) detector termed DyFrDet, which is able to precisely localize the small object by dynamically suppressing the background distractions in frequency domain. Specifically, we propose a Dynamic Frequency-aware Feature Pyramid Network (DyFrFPN) to adaptively suppress low-frequency redundancy and excessive high-frequency noises. The DyFrFPN transforms the hierarchical features into frequency domain representation, and introduces a Dynamic Band Predictor (DBP) to preserve the discriminative components for small object identification. Afterwards, we present a novel Label Disambiguation Module (LDM), which leverages probabilistic distributions to explicitly model and alleviate the inherent ambiguity of target labels, yielding efficient improvement in localization precision of the small objects with low-resolution. Extensive experiments demonstrate that DyFrDet achieves state-of-the-art performance across multiple benchmarks, indicating its effectiveness and robustness in various challenging scenarios.  Our code is available at \url{https://github.com/ManOfStory/DyFrDet}.
\end{abstract}

%%
%% The code below is generated by the tool at http://dl.acm.org/ccs.cfm.
%% Please copy and paste the code instead of the example below.
%%


%%
%% Keywords. The author(s) should pick words that accurately describe
%% the work being presented. Separate the keywords with commas.
%% A "teaser" image appears between the author and affiliation
%% information and the body of the document, and typically spans the
%% page.

%%
%% This command processes the author and affiliation and title
%% information and builds the first part of the formatted document.
\maketitle

\section{Introduction}
Small Object Detection (SOD), as a critical subtask of generic object detection, aims to accurately localize and recognize objects with limited size. SOD plays an essential role in a wide range of real-world applications, including unmanned aerial vehicle (UAV) surveillance, remote sensing localization, scene monitoring, obstacle avoidance, and autonomous driving. However, due to the lack of visual cues and complex scenarios, directly applying generic object detectors~\cite{girshick2014rich,ren2016faster,liu2016ssd,lin2017focal,carion2020end,zhu2021deformable} to small object detection often leads to significant performance degradation.

\begin{figure}[t]
\centering
\includegraphics[width=\linewidth]{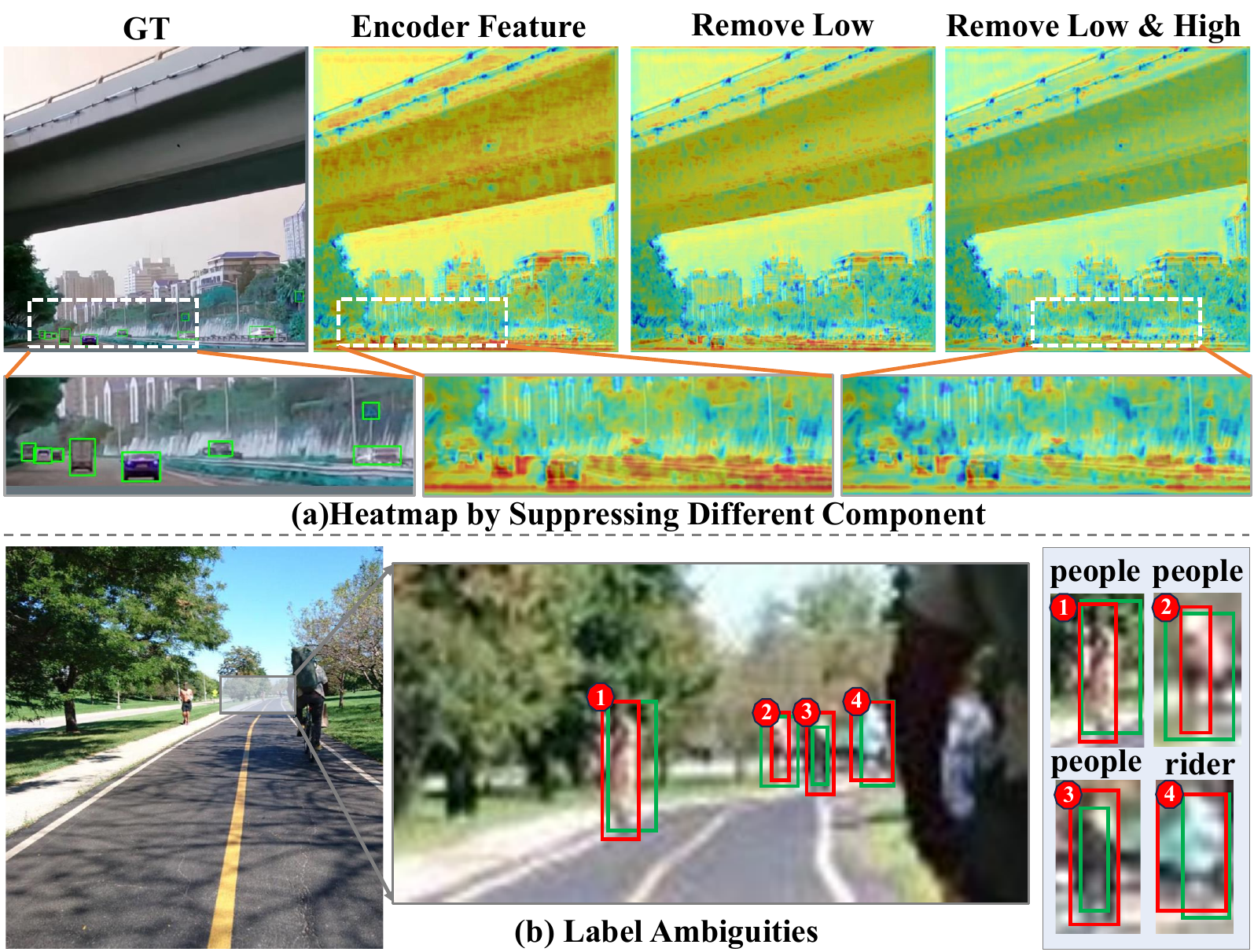}
\caption{
(a) Visualization of the heatmap using different frequency suppression settings. ``GT'' denotes ground-truth annotations. ``Encoder Feature'' is extracted from CFINet~\cite{CFINet}. ``Remove Low'' applies the low-frequency suppression strategy from HS-FPN~\cite{HS-FPN}. ``Remove Low $\&$ High'' further suppresses both low and high frequency components. (b) Label ambiguities in small targets. The green boxes indicate the ground truth annotations, whereas the red boxes are closer to the actual object location. Please zoom in for details.}
\label{fig:motivation}
% \vspace{-2mm}
\end{figure}

From a technical viewpoint, contemporary small object detection frameworks \cite{tan2020efficientdet,yang2022querydet,CFINet,10518058,sun2025set} predominantly make efforts on constructing semantically discriminative features to alleviate the deficiency of feature representation inherent to small objects. Although these approaches yield significant improvements in general small object detection, they still struggle to maintain robustness in highly cluttered complex scenarios.  Primarily, due to the diminutive spatial footprint, small objects are highly susceptible to interference from complex background textures and fine-grained noise, which often leads to false positives or missed detections. Some advanced approaches like HS-FPN \cite{HS-FPN} introduce a Discrete Cosine Transform (DCT) based filtering strategy to remove the low-frequency components, aiming to encourage the model to focus on the attentive small objects rather than large-scale distractors. However, the high-frequency distractors may still hamper the localization capability. Besides, the low resolution of small objects often leads to blurred boundaries, which inevitably introduces label ambiguity to supervise the model training, making it challenging to achieve accurate object localization.

To address these issues, in this paper, we propose a novel dynamic frequency suppressive detector termed DyFrDet, which is able to precisely localize the small object by dynamically suppressing the background distractions in frequency domain. As shown in Fig. \ref{fig:motivation}(a), the encoder features are typically noisy in complex local regions. If the low-frequency redundancies and high-frequency noises can be efficiently suppressed, the detection performance can be greatly boosted. To achieve this, DyFrDet introduces a Dynamic Frequency-aware Feature Pyramid Network (DyFrFPN) to transform the hierarchical pyramid features into different frequency components. A Dynamic Band Predictor (DBP) is designed to control the frequency representation, allowing the detector to suppress both low-frequency redundancies and high-frequency noises simultaneously. Meanwhile, as the low-resolution small objects often introduce label ambiguity, we propose a Label Disambiguation Module (LDM) to alleviate the performance degradation due to the unreliable model training under ambiguous annotations. The small object's coordinate regression is tackled as a position-aware distributional prediction task. By paying more attention to the high confidence samples and restraining the ambiguous samples, DyFrFPN can predict the small object's localization more precisely. Experimental results on several popular benchmarks validate the effectiveness of our proposed method. 

In conclusion, the main contributions of this paper are summarized as follows:
\begin{itemize}
    \item We propose DyFrDet, which introduces Dynamic Frequency-aware Feature Pyramid Network (DyFrFPN) to decompose the hierarchical pyramid features into different frequency components, and suppresses both low-frequency redundancies and high-frequency noises simultaneously for small object detection.
  
    \item We introduce a Label Disambiguation Module (LDM), which pays more attention to the high confidence samples and restrains the ambiguous samples, leading to more accurate bounding box estimation.
      
    \item Extensive experiments on several popular datasets verify the effectiveness and robustness of DyFrDet in handling extremely small objects under both densely and sparsely distributed scenarios.

\end{itemize}

\section{Related Work}
\subsection{Small Object Detection}
Recent advances in small object detection can be broadly categorized into three directions: data-oriented, architecture-oriented, and feature-oriented methods.
For data-oriented approaches, \cite{kisantal2019augmentation,chen2019rrnet} improve performance through data augmentation, while \cite{xu2021dot,xu2022detecting,xu2022rfla,CFINet} optimize the label assignment process during training, alleviating the scarcity of positive samples for small objects. These methods enhance supervision from the training perspective, thereby improving detection performance.
For architecture-oriented methods, FPN~\cite{lin2017feature} introduces a feature pyramid that leverages top-down aggregated multi-scale features for detection. QueryDet~\cite{yang2022querydet} proposes a cascade architecture and designs a Cascade Sparse Query (CSQ) mechanism to further boost performance. Other works~\cite{li2019scale,ghiasi2019fpn,tan2020efficientdet,qiao2021detectors, yao2026uaglnet, yao2024hierarchical} also explore advanced architectural designs, achieving improved performance through better multi-scale representation and feature utilization.
For feature-oriented methods, \cite{wu2020self,kim2021robust} adopt similarity learning to enhance representations by leveraging similar small objects. HS-FPN~\cite{HS-FPN} introduces a high-pass filter to suppress low-frequency components, encouraging the network to focus more on small objects. In addition, Sun \textit{et al.}~\cite{sun2025set} observe that high-frequency noise can degrade small object features and propose Spectral Enhancement (SET), which employs a heterogeneous architecture for foreground and background feature refinement.

Different from previous works, our method dynamically attenuates both low-frequency redundancy and high-frequency noise in a channel-wise manner, enabling more flexible and adaptive frequency modulation.

\subsection{Frequency Representation}
As an important tool in image processing, frequency domain analysis is widely employed in deep learning research. 
For feature modeling, Chi \textit{et al.}~\cite{chi2020fast} propose Fast Fourier Convolution (FFC), which extends conventional convolution by incorporating frequency-domain modeling, enabling non-local receptive fields and cross-scale feature interactions. 
Chen \textit{et al.}~\cite{chen2025frequency} introduce FDAM, which integrates dynamic high-pass and low-pass filters into the network architecture to address a key limitation of Vision Transformers (ViTs), namely their inherent low-pass filtering effect that leads to frequency attenuation and loss of fine-grained details. 
Qin \textit{et al.}~\cite{qin2021fcanet} show that global average pooling (GAP) is a special case of the discrete cosine transform (DCT), and further generalize channel attention mechanisms in the frequency domain via the proposed FcaNet, achieving strong performance on benchmarks such as ImageNet~\cite{deng2009imagenet} and COCO~\cite{lin2014microsoft}. 
SpectFormer~\cite{patro2025spectformer} introduces a hybrid architecture that combines spectral layers with multi-head self-attention, enabling the model to jointly capture global frequency representations and spatial dependencies. 
FSEL~\cite{sun2024frequency} incorporates frequency-domain transformations to alleviate the sensitivity and locality limitations of spatial features in camouflaged object detection.

In this paper, we jointly leverage frequency-domain and spatial features to better distinguish small objects from the background, dynamically suppressing irrelevant distractions and noise in feature representations.

\begin{figure*}[t]
\centering
\includegraphics[width=0.95\textwidth]{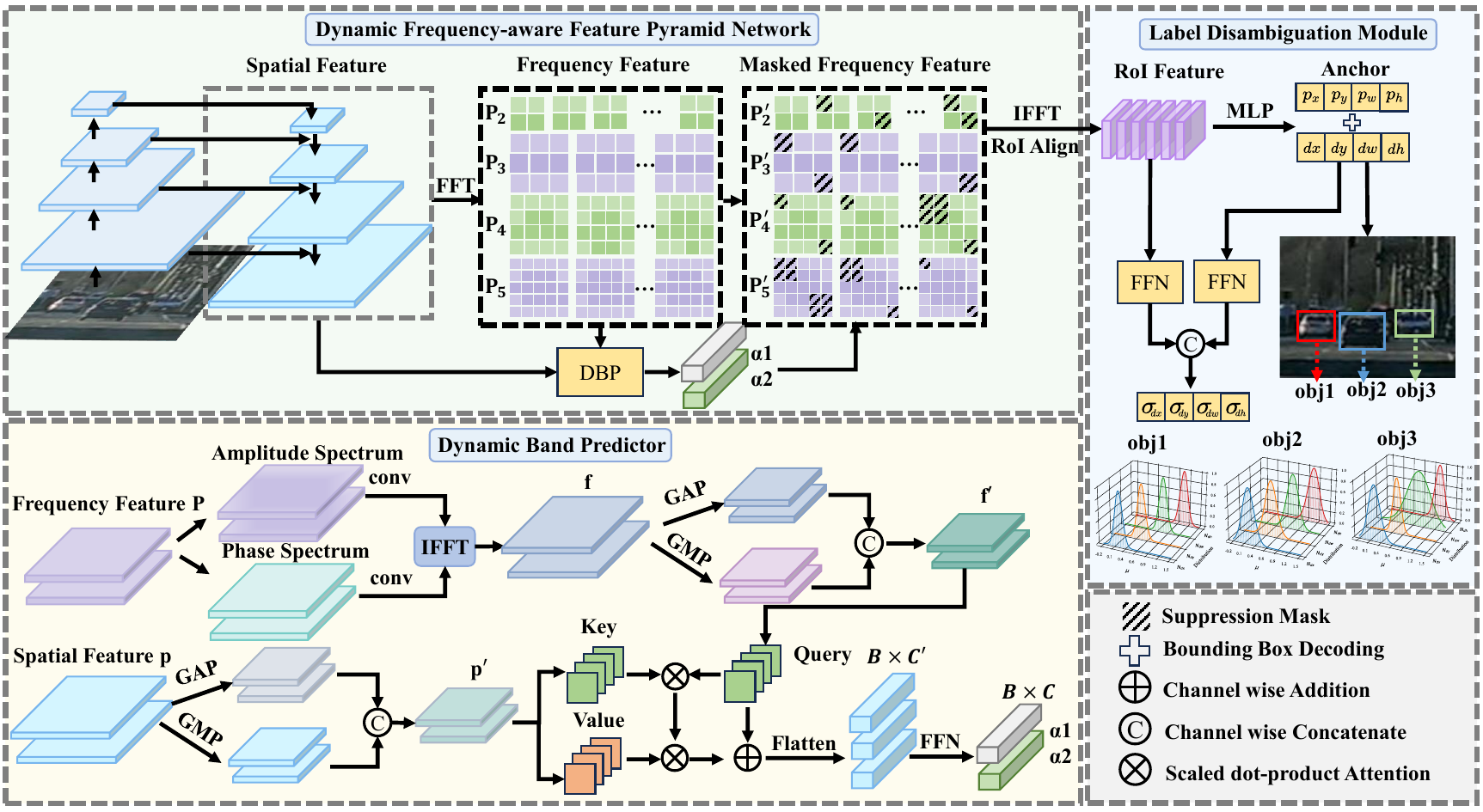}
\caption{The overall architecture of DyFrDet consists of two main components: Dynamic Frequency-aware Feature Pyramid Network (DyFrFPN) and Label Disambiguation Module (LDM). DyFrFPN integrates a Dynamic Band Predictor (DBP), which adaptively predicts two channel-wise frequency thresholds to suppress irrelevant components. The masked features are then passed to the LDM to enhance the final localization performance.
}
\label{fig:framework}
% \vspace{-2mm}
\end{figure*}

\subsection{Label Disambiguation}
Label ambiguity is often caused by human annotation bias or low-quality images.
This issue becomes especially severe in small object detection, where the distinction between object and background is often unclear.
Most existing works~\cite{9423558, ye2022joint, zhang2024mr, huang2024deim} treat bounding box regression as a deterministic prediction problem. However, such approaches exhibit significant limitations when faced with imprecise or ambiguous object boundaries.
Recently, modeling bounding boxes as probability distributions rather than fixed coordinates has attracted increasing attention across various domains, including visual tracking under uncertain or adverse conditions. For instance, UAST~\cite{zhang2022uast} discretizes localization outputs into probability distributions, while UMDATrack~\cite{yao2025umdatrack} and UncTrack~\cite{10967033} introduce uncertainty-aware tracking frameworks to improve localization reliability under challenging conditions. This probabilistic formulation has demonstrated strong capabilities in resolving label ambiguity and enhancing localization robustness under complex or uncertain conditions. However, to the best of our knowledge, the distributional bounding box regression for small object detection has not been carefully explored. In this work, we demonstrate the potential of our approach to mitigate label ambiguity and better handle complex scenarios in small object detection(SOD).

\section{Methodology}
In this section, we present the overall architecture of the proposed DyFrDet. As illustrated in Fig.~\ref{fig:framework}, DyFrDet consists of a  Dynamic Frequency-aware Feature Pyramid Network (DyFrFPN) and a Label Disambiguation Module(LDM). The DyFrFPN transforms pyramid features into frequency domain and employs a Dynamic Band Predictor (DBP) to adaptively suppress irrelevant frequency components. The LDM enforces the detector to focus on the high confidence samples and restrain the ambiguous samples to facilitate small object localization.

\subsection{Introduction to DyFrFPN}
Given an input image $\mathbf{I}_{t}$, we first pass it to a ResNet-50 backbone to extract multi-scale feature maps $\{\mathbf{c}_{2}, \mathbf{c}_{3}, \mathbf{c}_{4}, \mathbf{c}_{5}\}$, whose spatial resolutions are reduced by the factor of $\{4, 8, 16, 32\}$ and are passed through the proposed DyFrFPN. Specifically, the backbone feature maps are processed by a standard FPN to obtain the top-down aggregated multi-scale features $\{\mathbf{p}_{2}, \mathbf{p}_{3}, \mathbf{p}_{4}, \mathbf{p}_{5}\}$. These features are then transformed into the frequency domain using the Fast Fourier Transform (FFT), resulting in frequency representations $\{\mathbf{P}_{2}, \mathbf{P}_{3}, \mathbf{P}_{4}, \mathbf{P}_{5}\}$. The FFT is computed as follows:
\begin{equation}\label{eq:P_FFT}
P_i(u,v) = \sum_{x,y} p_i(x,y) \cdot e^{-j 2\pi \left( \frac{x}{H}u + \frac{y}{W}v \right)}, \quad j^2 = -1.
\end{equation}
where $i \in \{2,3,4,5\}$ and $p_i(x, y)$ denotes the spatial domain feature at location $(x, y)$ in level $i$, $P_i(u, v)$ is its corresponding frequency representation at coordinate $(u, v)$. 

This transformation enables the decomposition of features into different frequency components: the low-frequency signals in the top-left region of the spectrum and the high-frequency ones in the bottom-right. Since the multi-scale features differ only in spatial resolution while sharing the same processing operations, we omit the scale index $i$ in the following discussion for brevity.

\noindent\textbf{Dynamic Band Predictor.} Instead of applying uniform suppression across all channels, we predict the dynamic band range for each channel to conduct channel-wise precise denoising. To achieve this, we send frequency-domain representation $\mathbf{P} \in \mathbb{R}^{C \times H \times W}$ and the corresponding spatial-domain feature $\mathbf{p} \in \mathbb{R}^{C \times H \times W}$ into DBP to estimate the suppression thresholds $[\alpha_1, \alpha_2] \in \mathbb{R}^{C}$. Particularly, the complex-valued frequency feature $\mathbf{P}$ is first decomposed into the amplitude spectrum $\mathcal{A}$ and phase spectrum $\Phi$, which can be computed as follows:
\begin{equation}
\begin{aligned}
\mathcal{A}(u, v) &= \sqrt{\operatorname{Re}(P(u, v))^2 + \operatorname{Im}(P(u, v))^2}, \\
\Phi(u, v) &= \arctan\left( \frac{\operatorname{Im}(P(u, v))}{\operatorname{Re}(P(u, v))} \right),
\end{aligned}
\label{eq:amp_phase}
\end{equation}
where $\operatorname{Re}(\cdot)$ and $\operatorname{Im}(\cdot)$ denote the real and imaginary parts of the complex-valued feature, respectively.
After obtaining the amplitude spectrum $\mathcal{A}$ and phase spectrum $\Phi$, we pass them through a convolutional block to aggregate the frequency representations. The aggregated components are then transformed back into the spatial domain via the Inverse Fast Fourier Transform (IFFT). The complete process is formulated as:
\begin{equation}
\begin{aligned}
\mathcal{A}^{\prime}(u, v) &= \mathrm{ConvBlock}(\mathcal{A}(u, v)), \\
\Phi^{\prime}(u, v) &= \mathrm{ConvBlock}(\Phi(u, v)), \\
\mathbf{f} =\mathcal{F}^{-1}&\left[ \mathcal{A}^{\prime}(u, v) \cdot \exp({j \Phi^{\prime}(u, v)} \right)],\\
\end{aligned}
\label{eq:reconstruction}
\end{equation}
where $\mathcal{A}^{\prime}\in \mathbb{R}^{C\times H\times W}$ and $\Phi^{\prime}\in \mathbb{R}^{C\times H\times W}$ denote the amplitude and phase components, respectively.
Afterwards, we compress the spatial information of the feature maps using global pooling operations. We apply Global Max Pooling (GMP) and Global Average Pooling (GAP) to obtain compact feature representations, which can be given by:

% By capturing both local salient responses and global contextual distributions, the model can more effectively encode multi-scale semantic information.

\begin{equation}
\begin{aligned}
\mathbf{f}^\prime &= \operatorname{Concat}(\operatorname{GMP}(\mathbf{f}),\ \operatorname{GAP}(\mathbf{f})), \\
\mathbf{p}^\prime &= \operatorname{Concat}(\operatorname{GMP}(\mathbf{p}),\ \operatorname{GAP}(\mathbf{p})),
\end{aligned}
\end{equation}
where $\mathbf{f}^\prime\in \mathbb{R}^{2C\times H^{\prime}W^{\prime}}$ and $\mathbf{p}^\prime \in \mathbb{R}^{2C\times H^{\prime}W^{\prime}}$, we further project them into query, key, and value embeddings for attention computation as follows:
\begin{equation}
Q,\ K,\ V = [W_q, W_k, W_v]^T \cdot [\mathbf{p}^\prime,\ \mathbf{f}^\prime,\ \mathbf{f}^\prime],
\end{equation}
where $W_q$, $W_k$, $W_v \in \mathbb{R}^{2C\times d}$ are learnable linear projection matrices. $Q,K,V \in \mathbb{R}^{d\times H^{\prime}W^{\prime}}$ is the query, key and value embeddings.
To adaptively suppress the irrelevant noises, we predict the suppression thresholds $\alpha_1,\alpha_2 \in {R}^{C}$ through an attention mechanism followed by a feed-forward network (FFN). To ensure positivity and enable flexible scaling, we adopt an exponential parameterization as follows: 

\begin{equation}\label{eq:alpha}
\begin{aligned}
d\alpha &= \mathrm{FFN}\left( \mathbf{p}^\prime + \mathrm{softmax}\left( \frac{QK^\top}{\sqrt{d}} \right) V \right), \\
&\alpha_1 = \alpha_{l} \cdot e^{d\alpha_1}, \quad 
\alpha_2 = \alpha_{h} \cdot e^{d\alpha_2},
\end{aligned}
\end{equation}
where $\alpha_1, \alpha_2 \in \mathbb{R}^{C}$ denote the predicted lower and upper suppression thresholds, respectively. $\alpha_{l}, \alpha_{h} \in \mathbb{R}$ denote the predefined suppression thresholds. Based on the predicted $\alpha_1$ and $\alpha_2$, we define a spatial mask as follows:
\begin{equation}\label{eq:Mask} 
\mathcal{M}(x, y, j) = 
\begin{cases}
0, & \text{if } x < \alpha_1^{j} \cdot W \text{ and } y < \alpha_1^{j} \cdot H \\
0, & \text{if } x > \alpha_2^{j} \cdot W \text{ and } y > \alpha_2^{j} \cdot H \\
1, & \text{otherwise},
\end{cases}
\end{equation}
where $\alpha_1^j$ and $\alpha_2^j$ represent the lower and upper suppression thresholds of the $j$-th channel. The pixels within the top-left region (i.e., low-frequency redundant areas) and the bottom-right region (i.e., high-frequency noise areas) are masked out (set to 0), while the remaining spatial locations are retained (set to 1). 
By applying the predicted $\mathcal{M}$ to the frequency feature map $\mathbf{P}$, we obtain the noise suppressed feature $\mathbf{P}^{\prime}$. Which can be given by:
\begin{equation}\label{eq:PMask} 
\mathbf{P}^{\prime} = \mathbf{P} - \beta \cdot \left( \mathbf{P} \odot (1 - \mathcal{M}) \right),
\end{equation}
where $\odot$ denotes element-wise multiplication, and $\beta$ is a hyperparameter that controls the suppression rate. Finally, we convert the suppressed frequency-domain features back to the spatial domain using the Inverse Fast Fourier Transform (IFFT), yielding the enhanced multi-scale features $\{\mathbf{p^*_{2}}, \mathbf{p^*_{3}}, \mathbf{p^*_{4}}, \mathbf{p^*_{5}}\}$.

%Since different channels encode different features, this approach allows us to suppress irrelevant redundancy more precisely.

\subsection{Label Disambiguation Module}
\begin{figure}[t]
\centering
\includegraphics[width=0.95\linewidth]{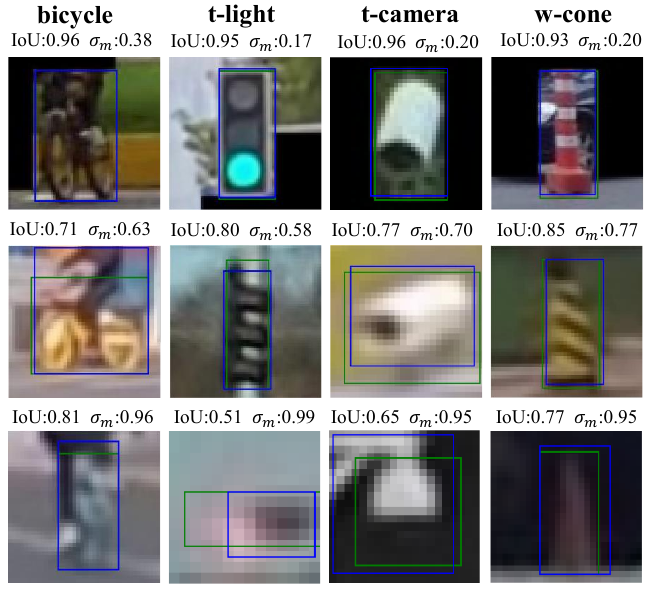}
\caption{
Visualization of different small objects under varying $\sigma_{m}$ values. The green boxes indicate ground-truth annotations, and blue boxes denote predicted bounding boxes. From top to bottom, the value of $\sigma_{m}$ gradually increases. The full names of class abbreviations are as follows: t-light (traffic-light), t-camera (traffic-camera), w-cone (warning cone).}
\label{fig:visualize_sigma}
% \vspace{-2mm}
\end{figure}

To precisely localize targets and mitigate the negative impact of label ambiguities, we introduce a Label Disambiguation Module~(LDM). Let $P = (P_x, P_y, P_w, P_h)$ denote the proposal box, where $(P_x, P_y)$ represents the center coordinates and $(P_w, P_h)$ denotes the size of the proposal, respectively. Similarly, $G = (G_x, G_y, G_w, G_h)$ represent the ground-truth bounding box. The bounding box regression process can be formulated as:
\begin{equation}
\begin{aligned}
t_x &= (G_x - P_x)/P_w,   &\quad \hat{G}_x &= P_w d_x + P_x, \\
t_y &= (G_y - P_y)/P_h,   &\quad \hat{G}_y &= P_h d_y + P_y, \\
t_w &= \log(G_w/P_w),     &\quad \hat{G}_w &= P_w \exp(d_w), \\
t_h &= \log(G_h/P_h),     &\quad \hat{G}_h &= P_h \exp(d_h),
\end{aligned}
\end{equation}
where $\mathcal{D} = \{d_x, d_y, d_w, d_h\}$ denotes the predicted offsets used to transform the proposal $P$ to the estimated bounding box $\hat{G}$. $\mathcal{T} = \{t_x, t_y, t_w, t_h\}$ represents the targets offsets corresponding to the ground truth box $G$ and anchor $P$. In the following discussion, we denote $\mathcal{D}$ and $\mathcal{T}$ by $\mathbf{x}$ and $\mathbf{x}_{\text{gt}}$ for simplicity.
Different from the traditional deterministic coordinate regression methods~\cite{ren2016faster,cai2018cascade,yao2025umdatrack} that minimize L1 loss between the predicted offsets $\mathbf{x}$ and the ground truth offset $\mathbf{x}_{\text{gt}}$, the proposed DyFrDet models them as a pair of Gaussian distribution and Dirac delta distribution, represented as:
\begin{equation}
\begin{aligned}
\mathbf{P}_{\boldsymbol{\theta}}(\mathbf{x}) = \mathcal{N}(\boldsymbol{\mu}_{\boldsymbol{\theta}}, \boldsymbol{\sigma}), 
    \quad 
    \mathbf{Q}_{\text{gt}}(\mathbf{x}) = \delta(\mathbf{x} - \mathbf{x}_{\text{gt}}),
\end{aligned}
\label{eq:distribution}
\end{equation}
where $\mathbf{P}_{\boldsymbol{\theta}}(\mathbf{x})$ denotes the predicted localization distribution, and $\mathbf{Q}_{\text{gt}}(\mathbf{x})$ represents the ground-truth distribution. Here $\boldsymbol{\mu}$ is the mean of the Gaussian distribution, and $\boldsymbol{\sigma}$ is the predicted covariance that determines the quality of the localization. Then the label quality can be reflected by the maximum value of $\boldsymbol{\sigma}$, which is denoted as $\sigma_{m}$. As shown in Fig.~\ref{fig:visualize_sigma}, the object boundaries become increasingly blurred as $\sigma_{m}$ increases from top to bottom. 
% When $\sigma_{m} > 0.9$, the label ambiguity becomes significant, which hinders accurate localization.

During training, we minimize the Kullback–Leibler (KL) divergence between the predicted distribution $\mathbf{P}_{\boldsymbol{\theta}}(\mathbf{x})$ and the ground-truth distribution $\mathbf{Q}_{\text{gt}}(\mathbf{x})$. The loss function is defined as:

\begin{equation}
\begin{aligned}
    \mathcal{L}_{\mathrm{LDM}} &= \mathrm{KL}(\mathbf{Q}_{\text{gt}}(\mathbf{x}) \parallel \mathbf{P}_{\boldsymbol{\theta}}(\mathbf{x})) \\
           &\propto -\mathbb{E}_{\mathbf{x} \sim \mathbf{Q}_{\text{gt}}} \left[ \log \mathbf{P}_{\boldsymbol{\theta}}(\mathbf{x}) \right]\\
           &\propto \frac{(\mathbf{x}_{\text{gt}} - \boldsymbol{\mu})(\mathbf{x}_{\text{gt}} - \boldsymbol{\mu})^\top}{ (\boldsymbol{\sigma}^\top \boldsymbol{\sigma})} + \log(\boldsymbol{\sigma}^\top \boldsymbol{\sigma}),
\end{aligned}
\label{eq:kl_loss}
\end{equation}
To mitigate the negative impact of these samples with large label ambiguities, we design a weighted function as follows:
\begin{equation}
\omega(\sigma_{m}) = 
\left\{
\begin{array}{ll}
1 & \text{if } \sigma_{m} \leq \rho \\
\varepsilon + (1 - \varepsilon)\left[1 - \left(\dfrac{\sigma_{m} - \rho}{1 - \rho} \right)^2 \right] & \text{if } \rho < \sigma_{m} < 1,
\end{array}
\right.
\end{equation}
where $\varepsilon$ and $\rho$ are hyperparameters to control the penalty range. When $\sigma_{m} \le \rho$, $\omega(\sigma_{m})$ remains 1, while when $\sigma_{m} \to 1$, $\omega(\sigma_{m})$ becomes closer to $\varepsilon$. By reweighting the loss term, we are able to reduce the detrimental effects of label ambiguity in training. Finally, the overall loss function is defined as:
\begin{equation}
    \mathcal{L} = \gamma \mathcal{L}_{\text{IoU}} + \mathcal{L}_{\text{CLS}} + \omega(\sigma_{m})*\mathcal{L}_{\text{LDM}},
\end{equation}
where $\mathcal{L}_{\text{IoU}}$ denotes the IoU loss. $\mathcal{L}_{\text{CLS}}$ is the cross-entropy classification loss for candidate proposals. The coefficient $\gamma$ is a hyperparameter that balances the loss contribution.

\section{Experiment}
In this section, we first provide the implementation details of our method and comparisons with other state-of-the-art detectors on two popular benchmarks: \textbf{AI-TOD}~\cite{AI-TOD_2020_ICPR} and \textbf{SODA}~\cite{cheng2023towards}. SODA includes two subsets, \textbf{SODA-D} and \textbf{SODA-A}. Next, we conduct ablation studies to evaluate the effectiveness of our approach. Finally, we present qualitative visualizations to analyze the impact on feature representations.

\noindent\textbf{Implementation Details}
All experiments are conducted on a single NVIDIA RTX 3090 GPU. For AI-TOD, the patch size is fixed to $800 \times 800$, while for SODA-D and SODA-A, the patches are resized to $1200 \times 1200$. The hyperparameters $\alpha_{l}$ and $\alpha_{h}$ are set to 0.05 and 0.95. The suppression rate $\beta$ is set to 0.5. The weight function parameters $\varepsilon$ and $\rho$ are set to 0.5, 0.8, respectively. The loss weights $\gamma$ are set to 0.9. The model is trained for 36 epochs on AI-TOD and for 12 epochs on SODA-D and SODA-A.
To ensure that the suppressed frequency components correspond to meaningful distractors, the Dynamic Frequency Suppression Strategy is activated from the 24th epoch on AI-TOD and the 8th epoch on both SODA-D and SODA-A.

\subsection{Comparison with state-of-the-art}
\noindent\textbf{AI-TOD.}
The AI-TOD dataset contains 700,621 instances in 8 categories in 28,036 aerial images. Compared to existing object detection datasets in aerial images, the mean size of objects in AI-TOD is about 12.8 pixels, which is much smaller than others.

We report the results of our proposed DyFrDet on AI-TOD test set. As shown in Table \ref{tab:aitod_comparison}, DyFrDet achieves the best performance across all metrics. Compared with HS-FPN \cite{HS-FPN}, it improves (AP, AP$_{50}$, AP$_{75}$, AP$_{\text{vt}}$, AP$_{\text{t}}$, AP$_{\text{s}}$, AP$_{\text{m}}$) by (3.6\%, 1.5\%, 4.6\%, 3.1\%, 3.8\%, 3.4\%, 2.7\%), respectively.

\setlength{\tabcolsep}{1mm}
\begin{table*}[t]
\centering
\caption{Comparison with state-of-the-art detection methods on AI-TOD test set. All metrics follow the COCO-style evaluation protocol, including overall AP and performance across different scene types. $^{*}$ indicates methods using ResNet-50 as the backbone. \textbf{Bold} numbers indicate the best results.}
\resizebox{0.95\textwidth}{!}{
\begin{tabular}{l|c|c|ccccccc}
\hline
Method & Source & Backbone & AP & AP$_{50}$ & AP$_{75}$ & AP$_{\text{vt}}$ & AP$_{\text{t}}$ & AP$_{\text{s}}$ & AP$_{\text{m}}$ \\ \hline
\multicolumn{10}{c}{\textbf{Query-based Detectors}} \\\hline
DAB-DETR\cite{DBLP:conf/iclr/LiuLZYQSZZ22} & ICLR2022 & R50 & 4.9 & 16.0 & 1.7 & 1.7 & 3.6 & 7.0 & 18.0 \\
DAB-Deformable-DETR\cite{DBLP:conf/iclr/LiuLZYQSZZ22} & ICLR2022 & R50 & 16.5 & 42.6 & 9.9 & 7.9 & 15.2 & 23.8 & 31.9 \\
DINO-Defomable-DETR\cite{DBLP:conf/iclr/0097LL000NS23} & ICLR2023 & R50 & 23.2 & 56.6 & 15.4 & 9.9 & 23.1 & 29.3 & 37.6 \\
DINO-5scale w/SET \cite{sun2025set} & CVPR2025 & R50 & 26.6 & 57.1 & 20.8 & 13.2 & 27.1 & 31.5 & -- \\\hline
\multicolumn{10}{c}{\textbf{Multi-Stage Detectors}} \\\hline
Faster R-CNN \cite{ren2016faster} & TPAMI2017 & R50+FPN & 11.1 & 26.3 & 7.6 & 0.0 & 7.2 & 23.3 & 33.6 \\
Cascade R-CNN \cite{cai2018cascade} & CVPR2018 & R50+FPN & 13.8 & 30.8 & 10.5 & 0.0 & 10.5 & 25.5 & 36.6 \\
DetectorRS\cite{qiao2021detectors} & CVPR2021 & R50+FPN & 14.8 & 32.8 & 11.4 & 0.0 & 10.8 & 28.3 & 38.0 \\
QueryDet\cite{yang2022querydet}            & CVPR2022  & R50+FPN & 12.2 & 29.3 & 7.9 & 2.4 & 10.5 & 18.5 & 26.3 \\
CFINet\cite{CFINet} & ICCV2023 & R50+FPN & 24.7 & 53.9 & 18.6 & 11.7 & 26.4 & 28.1 & 32.2 \\
KLDet\cite{zhou2024kldet} & TGRS2024 & R50+FPN & 19.6 & 46.4 & 13.7 & 8.4 & 20.6 & 22.7 & 26.4 \\
RFLA\cite{xu2022rfla} & ECCV2022 & R50 w/SAC+FPN & 24.8 & 55.2 & 18.5 & 9.3 & 24.8 & 30.3 & 38.2 \\
DNTR\cite{10518058} & TGRS2024 & R50 w/SAC+FPN & 26.2 & 56.7 & 20.2 & 12.8 & 26.4 & 31.0 & 37.0 \\
SimD\cite{shi2024similarity} & IROS2024 & R50 w/SAC+FPN & 26.6 & 55.9 & 21.2 & 13.4 & 27.5 & 30.9 & 37.8 \\
% FBRT-YOLO-S\cite{xiao2025fbrt} & AAAI2025 & -- & 20.2 & 45.8 & -- & -- & -- &-- & --  \\ 
DetectorRS w/FIDP \cite{bian2025feature} & CVPR2025 & R50 w/SAC+FPN & 24.3 & 54.4 & 18.3 & 8.5 & 24.9 & 29.8 & -- \\
HS-FPN\cite{HS-FPN} & AAAI2025 & R50 w/SAC+FPN & 25.1 & 55.7 & 19.1 & 12.1 & 25.3 & 29.9  & 36.9 \\ 
\hline
\textbf{DyFrDet$^{*}$}           & --         & R50+DyFrFPN       & 26.0 & 55.3   & 19.6   & 13.6   & 27.3  & 29.0  & 34.5  \\
\textbf{DyFrDet}           & --         & R50 w/SAC+DyFrFPN & \textbf{28.7} & \textbf{57.2}   & \textbf{23.7}   & \textbf{15.2}   & \textbf{29.1}  & \textbf{33.3}  & \textbf{39.6}  \\ 
\hline
\end{tabular}
}
\label{tab:aitod_comparison}
\end{table*}

\begin{table}[htb]
\centering
\caption{Ablation study on AI-TOD test set. When DyFrFPN is disabled, it is replaced with a standard FPN. When LDM is removed, an FFN with L1 loss is used instead.}
\renewcommand{\arraystretch}{1.1}
\resizebox{0.75\linewidth}{!}{
\begin{tabular}{cc|ccccc}
\hline
DyFrFPN & LDM & AP & AP$_\text{vt}$ & AP$_\text{t}$ & AP$_\text{s}$ & AP$_\text{m}$\\
\hline
$\times$ & $\times$ & 23.6 & 11.9 & 24.8 & 27.0 & 30.7\\
$\checkmark$ & $\times$ & 24.9 & 12.3 & 26.2 & 28.7 & 33.3\\
$\times$ & $\checkmark$  & 24.4 & 12.0 & 25.6 & 27.3 & 32.5\\
$\checkmark$ & $\checkmark$ & \textbf{26.0} & \textbf{13.6} & \textbf{27.3} & \textbf{29.0} & \textbf{34.5}\\
\hline
\end{tabular}
}
% \vspace{-2mm}
\label{tab:ablation_soda_d}
\end{table}

\begin{table}[t]
\centering
\caption{Comparison of the static and dynamic frequency suppression strategies. ``Static'' denotes suppressing frequency components outside the fixed thresholds [$\alpha_1$, $\alpha_2$] during the FPN stage. ``Dynamic'' refers to the proposed DyFrFPN, which dynamically predicts $\alpha_1$ and $\alpha_2$ to suppress noise. All experiments are conducted on AI-TOD test set.}
\renewcommand{\arraystretch}{1.1} % 增加行间距
\resizebox{0.85\linewidth}{!}{
\begin{tabular}{c|cc|ccccc}
\hline
Strategy & $\alpha_1$ & $\alpha_2$ & AP & AP$_\text{vt}$ & AP$_\text{t}$ & AP$_\text{s}$ & AP$_\text{m}$\\
\hline
None   & -- & -- & 24.4 & 12.0 & 25.6 & 27.3 & 32.5\\
Static   & 0.05 & 1.00 & 24.9 & 12.1 & 26.4 & 28.3 & 32.1\\
Static & 0.05 & 0.95 & 25.3 & 13.1 & 26.7 & 28.8 & 32.8\\
Static & 0.10 & 0.90 & 24.8 & 12.3 & 26.1 & 27.7 & 33.3\\
Static & 0.15 & 0.85 & 24.7 & 11.8 & 26.0 & 28.0 & 32.5\\
Static & 0.20 & 0.80 & 25.1 & 11.9 & 26.6 & 28.5 & 32.4\\
Static & 0.25 & 0.75 & 24.7 & 12.8 & 25.8 & 28.8 & 32.6\\
Static & 0.30 & 0.70 & 25.0 & 11.5 & 26.2 & 28.3 & 32.3 \\
Dynamic  & -- & -- & \textbf{26.0} & \textbf{13.6} & \textbf{27.3} & \textbf{29.0} & \textbf{34.5} \\
\hline
\end{tabular}
}
% \vspace{-2mm}
\label{tab:strategy_ablation}
\end{table}

\noindent\textbf{SODA-A.}
The SODA-A dataset consists of 2,513 aerial images with 872,069 oriented bounding box annotations, focusing on small object detection. SODA-A has an average resolution of 4761 × 2777 pixels and contains approximately 347 instances per image, presenting a highly dense object distribution that challenges existing detection models in clustered scenes.

\setlength{\tabcolsep}{1mm} % 适当压缩列宽，避免溢出
\begin{table*}[t]
\centering
\caption{Comparison with oriented object detection methods on SODA-A test set. Metrics follow COCO-style evaluation. $^{*}$ indicates methods using ResNet-50 as the backbone. \textbf{Bold} numbers indicate the best results.}
\renewcommand{\arraystretch}{1.05}
\resizebox{0.95\textwidth}{!}{
\begin{tabular}{l|c|c|ccccccc}
\hline
Method & Source & Backbone & AP & AP$_{50}$ & AP$_{75}$ & AP$_\text{es}$ & AP$_\text{rs}$ & AP$_\text{gs}$ & AP$_{N}$ \\
\hline
\multicolumn{10}{c}{\textbf{One-stage Detectors}} \\\hline
Rotated RetinaNet\cite{lin2017focal}  & ICCV2017 & R50 + FPN & 26.8 & 63.4 & 16.2 & 9.1  & 22.0 & 35.4 & 28.2 \\

Oriented RepPoints\cite{li2022oriented} & CVPR2022 & R50 + FPN & 26.3 & 58.8 & 19.0 & 9.4  & 22.6 & 32.4 & 28.5 \\
DHRec\cite{nie2022multi} & TPAMI2022 & R50 + FPN & 30.1 & 68.8 & 19.8 & 10.6 & 24.6 & 40.3 & 34.6 \\
CFPT\cite{11010850} & TGRS2025 & R50+CFPT & 25.9 & 63.3 & 14.3 & 9.0 & 21.2 & 34.6 & 27.9 \\
LEGNet\cite{lu2025legnet} & ICCVW2025 & LWGNet + FPN & 29.6 & 58.7 & 26.4 & 10.4 & 26.0 & 39.6 & 32.0 \\
\hline
\multicolumn{10}{c}{\textbf{Two-stage Detectors}} \\\hline
Rotated Faster R-CNN\cite{ren2016faster}  & TPAMI2017 & R50 + FPN & 32.5 & 70.1 & 24.3 & 11.9 & 27.3 & 42.2 & 34.4 \\
Gliding Vertex\cite{xu2020gliding}   & TPAMI2020 & R50 + FPN & 31.7 & 70.8 & 22.6 & 11.7 & 27.0 & 41.1 & 33.8 \\
Oriented R-CNN\cite{xie2021oriented} & ICCV2021 & R50 + FPN & 34.4 & 70.7 & 28.6 & 12.5 & 28.6 & 44.5 & 36.7 \\
DODet\cite{cheng2022dual} & TGRS2022 & R50 + FPN & 31.6 & 68.1 & 23.4 & 11.3 & 26.3 & 41.0 & 33.5 \\
CFINet\cite{CFINet} &ICCV2023 & R50 + FPN & 34.4 & 73.1 & 26.1 & 13.5 & 29.3 & 44.0 & 35.9 \\
DecoupleNet\cite{lu2024decouple} & TGRS2024 & DecoupleNet + FPN & 36.6 & 71.3 & 33.3 & 12.2 & 31.0 & 47.7 & 40.2 \\ 
STD\cite{yu2024spatial} & AAAI2024 & R50+FPN & 30.3 & 67.4 & 21.2 & 12.4 & 26.6 & 36.7 & 30.7 \\
SR-TOD\cite{cao2024visible} & ECCV2024 & R50+FPN & 32.7 & 70.5 & 24.1 & 12.2 & 27.5 & 42.6 & 34.5 \\
GauCho\cite{Marques_2025_CVPR} & CVPR2025 & R50 + FPN & 33.2 & 70.1 & 25.0 & 9.9 & 27.8 & 44.9 & 36.4\\
Unc-SOD\cite{yuan2026unc} & TIP2026 & R50+FPN & 34.8 & \textbf{73.6} & 26.4 & 13.8 & 29.7 & 44.7 & 36.5 \\
\hline
\textbf{DyFrDet$^{*}$} & -- & R50 + DyFrFPN  & 36.0 & 73.3 & 30.1 & \textbf{13.8} & 30.6 & 46.0 & 38.0 \\
\textbf{DyFrDet} & -- & DecoupleNet + DyFrFPN  & \textbf{37.8} & 73.4 & \textbf{34.3} & 12.9 & \textbf{31.9} & \textbf{49.5} & \textbf{41.0} \\
\hline
\end{tabular}
}
%\vspace{1mm}

\label{tab:sodaa_comparison}
\end{table*}

Table~\ref{tab:sodaa_comparison} summarizes the performance on SODA-A benchmark. DyFrDet achieves state-of-the-art results with (37.8\%, 73.4\%, 34.3\%, 31.9\%, 49.5\%, 41.0\%) on (AP, AP$_{50}$, AP$_{75}$, AP$_\text{rs}$, AP$_\text{gs}$, AP$_\textbf{N}$), surpassing state-of-the-art GauCho~\cite{Marques_2025_CVPR} by (4.6\%, 3.3\%, 9.3\%, 3.0\%, 4.1\%, 4.6\%, 4.6\%) across all metrics.

\begin{table}[t]
\centering
\caption{Ablation study of different compositional inputs of DBP. ``FF'' denotes the method only using the frequency feature, ``SF'' only uses the spatial feature, and ``FF+SF'' uses both. Results are reported in terms of Average Precision (AP) on various object sizes.}
\label{tab:DBP_ablation}
\resizebox{0.7\linewidth}{!}{
\begin{tabular}{c|ccccc}
\hline
Strategy & AP & AP$_\text{vt}$ & AP$_\text{t}$ & AP$_\text{s}$ & AP$_\text{m}$ \\
\hline
FF & 24.8 & 11.9 & 26.2 & 28.2 & 32.7 \\
SF & 24.6 & 12.2 & 26.0 & 28.2 & 32.7 \\
\textbf{FF+SF} & \textbf{26.0} & \textbf{13.6} & \textbf{27.3} & \textbf{29.0} & \textbf{34.5} \\
\hline
\end{tabular}}
% \vspace{-2mm}
\end{table}

\begin{table}[t]
\centering
\caption{Ablation study on different suppression rates $\beta$ in Eq.~\ref{eq:PMask}. $\beta=0.0$ indicates no suppression, while $\beta=1.0$ denotes direct filtering.}
\renewcommand{\arraystretch}{1.1}
\resizebox{0.65\linewidth}{!}{
\begin{tabular}{c|ccccc}
\hline
$\beta$ & AP & AP$_\text{vt}$ & AP$_\text{t}$ & AP$_\text{s}$ & AP$_\text{m}$\\
\hline
0.00 & 24.4 & 12.0 & 25.6 & 27.3 & 32.5\\
0.25 & 25.0 & 12.5 & 26.1 & 29.0 & 32.9\\
0.50 & \textbf{26.0} & \textbf{13.6} & \textbf{27.3} & \textbf{29.0} & \textbf{34.5}\\
0.75 & 24.6 & 11.9 & 25.9 & 27.5 & 32.3\\
1.00 & 25.3 & 12.7 & 26.5 & 28.6 & 33.2\\
\hline
\end{tabular}}
% \vspace{-2mm}
\label{tab:beta_ablation}
\end{table}

% \begin{table}[tbp]
% \centering
% \footnotesize
% \setlength{\tabcolsep}{3pt}
% \renewcommand{\arraystretch}{1.1}

% \caption{Complexity comparison with existing methods.}
% \label{tab:efficiency}
% \resizebox{\linewidth}{!}{
% \begin{tabular}{l|ccc|c}
% \hline
% & HS-FPN~\cite{HS-FPN} & DNTR~\cite{10518058} & CFINet~\cite{CFINet} & DyFrDet$^*$ \\
% \hline
% GFLOPs      & 200.7 & 368.9 & 154.9 & 177.8 \\
% Params (M)  & 71.7  & 128.4 & 44.0  & 75.8  \\
% AP (\%)         & 29.6  & 29.6  & 30.7  & 31.7  \\
% \hline
% \end{tabular}
% }
% % \vspace{-2mm}
% \end{table}

\noindent\textbf{SODA-D.} 
The SODA-D dataset contains 24,828 high-resolution images with 278,433 annotated instances spanning 9 categories: \textit{people}, \textit{rider}, \textit{bicycle}, \textit{motor}, \textit{vehicle}, \textit{traffic sign}, \textit{traffic light}, \textit{traffic camera}, and \textit{warning cone}. It exhibits rich diversity in terms of locations, weather conditions, period, camera viewpoints, and traffic scenarios. With an average image resolution of 3407 × 2470, the dataset is particularly well-suited for detecting small and tiny objects in complex environments.

We report the performance of DyFrDet on SODA-D test set, as summarized in Table~\ref{tab:sodad_comparison}. Our DyFrDet achieves state-of-the-art performance across all metrics, with an average AP of 31.3\%. 
Compared with HS-FPN\cite{HS-FPN}, DyFrDet yields consistent improvements of (1.7\%, 5.3\%, 0.1\%, 1.5\%, 1.4\%, 2.0\%, 0.9\%) in terms of (AP, AP$_{50}$, AP$_{75}$, AP$_\text{es}$, AP$_\text{rs}$, AP$_\text{gs}$, AP$_\text{N}$).

\setlength{\tabcolsep}{1mm} % 适当压缩列宽，避免溢出
\begin{table*}[t]
\centering
\caption{Comparison with state-of-the-art detection approaches on SODA-D test set. All metrics follow COCO-style evaluation, including overall AP and performance across different scene types. \textbf{Bold} numbers indicate the best results. }
\renewcommand{\arraystretch}{1.05}
\resizebox{0.80\textwidth}{!}{
\begin{tabular}{l|c|c|ccccccc}
\hline
Method & Source & Backbone & AP & AP$_{50}$ & AP$_{75}$ & AP$_{\text{es}}$ & AP$_{\text{rs}}$ & AP$_{\text{gs}}$ & AP$_N$ \\
\hline
\multicolumn{10}{c}{\textbf{One-stage Detectors}} \\\hline
RetinaNet\cite{lin2017focal}& ICCV2017 & R50+FPN & 28.2 & 57.6 & 23.7 & 11.9 & 25.2 & 34.1 & 44.2 \\
FCOS\cite{tian2019fcos} & ICCV2019 & R50+FPN & 23.9 & 49.5 & 19.9 & 6.9  & 19.4 & 30.9 & 40.9 \\
ATSS\cite{zhang2020bridging} & CVPR2020 & R50+FPN & 26.8 & 55.6 & 22.1 & 11.7 & 23.9 & 32.2 & 41.3 \\
DyHead\cite{dai2021dynamic} & CVPR2021 & R50+FPN  & 27.5 & 56.1 & 23.2 & 12.4 & 24.4 & 33.0 & 41.9 \\
KLDet\cite{zhou2024kldet} & TGRS2024 & R50+FPN & 25.9 & 53.8 & 21.4 & 10.7 & 22.2 & 31.9 & 41.6 \\
CFPT\cite{11010850} & TGRS2025 & R50+CFPT & 27.5 & 54.5 & 23.8 & 7.1 & 22.4 & 36.2 & 45.9 \\
\hline
\multicolumn{10}{c}{\textbf{Two-stage Detectors}} \\\hline
Faster R-CNN\cite{ren2016faster} & TPAMI2017 & R50+FPN & 28.9 & 59.7 & 24.2 & 13.9 & 25.6 & 34.3 & 43.2 \\
Cascade RPN\cite{vu2019cascade} & NIPS2019 & R50+FPN & 29.1 & 56.5 & 25.9 & 12.5 & 25.5 & 35.4 & 44.7 \\
KL\cite{he2019bounding} & CVPR2019 & R50+FPN & 29.4 & 59.2 & 24.7 & 14.0 & 26.3 & 36.0 & 44.2 \\
RFLA\cite{xu2022rfla} & ECCV2022 & R50+FPN & 29.7 & 60.2 & 25.2 & 13.2 & 26.9 & 35.4 & 44.6 \\
CFINet\cite{CFINet} & ICCV2023 & R50+FPN & 30.7 & 60.8 & 26.7 & 14.7 & 27.8 & 36.4 & 44.6 \\ 
DNTR\cite{10518058} & TGRS2024 & R50 w/SAC+RFP & 29.6 & 57.8 & 26.5 & 13.1 & 26.7 & 35.5 & 43.4 \\
SR-TOD\cite{cao2024visible} & ECCV2024 &  R50+FPN &29.3 & 60.0 & 24.5 & 13.8 & 26.0 & 35.6 & 43.4 \\
HS-FPN\cite{HS-FPN} & AAAI2025 & R50 + FPN & 29.6 & 56.8 & 26.7 & 13.6 & 26.4 & 35.3 & 45.3 \\
\hline
\textbf{DyFrDet$^{*}$} & -- &  R50+DyFrFPN & \textbf{31.3} & \textbf{62.1} & \textbf{26.8} & \textbf{15.1} & \textbf{27.8} & \textbf{37.3} & \textbf{46.2} \\
\hline
\end{tabular}
}
% \vspace{1mm}
\label{tab:sodad_comparison}
\end{table*}

\subsection{Ablation Study}
\noindent\textbf{Ablation of Different Variations.} To evaluate the effectiveness of each component, we selectively ablate DyFrFPN and LDM from DyFrDet. When DyFrFPN is removed, it is replaced with a standard FPN. Similarly, when LDM is removed, it is replaced with an FFN head trained using the L1 loss.

\setlength{\tabcolsep}{4pt} % 可微调压缩列间距
\renewcommand{\arraystretch}{1.2} % 控制行高，提升可读性

As shown in Table~\ref{tab:ablation_soda_d}, the baseline detector without DyFrFPN and LDM achieves (23.6\%, 11.9\%, 24.8\%, 27.0\%, 30.7\%) in terms of (AP, AP$_\text{vt}$, AP$_\text{t}$, AP$_\text{s}$, AP$_\text{m}$), respectively. When DyFrFPN is introduced, the performance improves to (24.9\%, 12.3\%, 26.2\%, 28.7\%, 33.3\%), demonstrating its ability to effectively suppress low-frequency redundancies and high-frequency noises. Additionally, incorporating LDM into the baseline also leads to performance gains, achieving (24.4\%, 12.0\%, 25.6\%, 27.3\%, 32.5\%) for (AP, AP$_\text{vt}$, AP$_\text{t}$, AP$_\text{s}$, AP$_\text{m}$), respectively. Furthermore, combining both DyFrFPN and LDM yields the best performance, with improvements of (2.4\%, 1.7\%, 2.5\%, 2.0\%, 3.8\%) over the baseline. These results validate the effectiveness of the proposed DyFrFPN and LDM modules.

\noindent\textbf{Ablation of Different Suppression Strategies.}
To verify the effectiveness of the proposed Dynamic Frequency Suppression Strategy, we analyze the performance gains under different frequency suppression strategies. In our experiments, ``None'' denotes that no frequency components are suppressed. ``Static'' with various settings of $\alpha_1$ and $\alpha_2$ refers to the strategy that statically suppresses frequency components outside the range $[\alpha_1, \alpha_2]$.

As shown in Table~\ref{tab:strategy_ablation}, when following the setting of HS-FPN~\cite{HS-FPN}, where only low-frequency components are suppressed by setting $\alpha_1 = 0.05$ and $\alpha_2 = 1.00$, the performance improves (0.5\%, 0.1\%, 0.8\%, 1.0\%) in terms of (AP, AP$_\text{vt}$, AP$_\text{t}$, AP$_\text{s}$), respectively.
When high-frequency suppression is also introduced (e.g., setting $\alpha_1 = 0.05$ and $\alpha_2 = 0.95$), the performance further improves (0.4\%, 1.0\%, 0.3\%, 0.5\%, 0.7\%) over the previous setting in terms of (AP, AP$_\text{vt}$, AP$_\text{t}$, AP$_\text{s}$, AP$_\text{m}$). These experiments validate the importance of suppressing high-frequency noise.
Moreover, the dynamic strategy yields the best overall performance, reaching (26.0\%, 13.6\%, 27.3\%, 29.0\%, 34.5\%) on (AP, AP$_\text{vt}$, AP$_\text{t}$, AP$_\text{s}$, AP$_\text{m}$), which demonstrates the effectiveness of our proposed suppression strategy.

\noindent\textbf{Band Predictor Settings.}
We perform an ablation study to analyze the impact of different input feature compositions for the Dynamic Band Predictor (DBP), as reported in Table~\ref{tab:DBP_ablation}. Specifically, we investigate three variants that utilize frequency features (FF) alone, spatial features (SF) alone, and the combination of both (FF+SF). When using either individual branch, FF and SF achieve comparable performance, obtaining overall AP scores of 24.8 and 24.6, respectively, which indicates that both frequency-domain representations and spatial cues provide valuable information for band prediction. By jointly incorporating FF and SF, DBP achieves consistent improvements across all object scales and obtains the highest overall AP of 26.0, validating the necessity of each branch and the complementary roles of frequency and spatial information.

\begin{figure}[t]
\centering
\includegraphics[width=\linewidth]{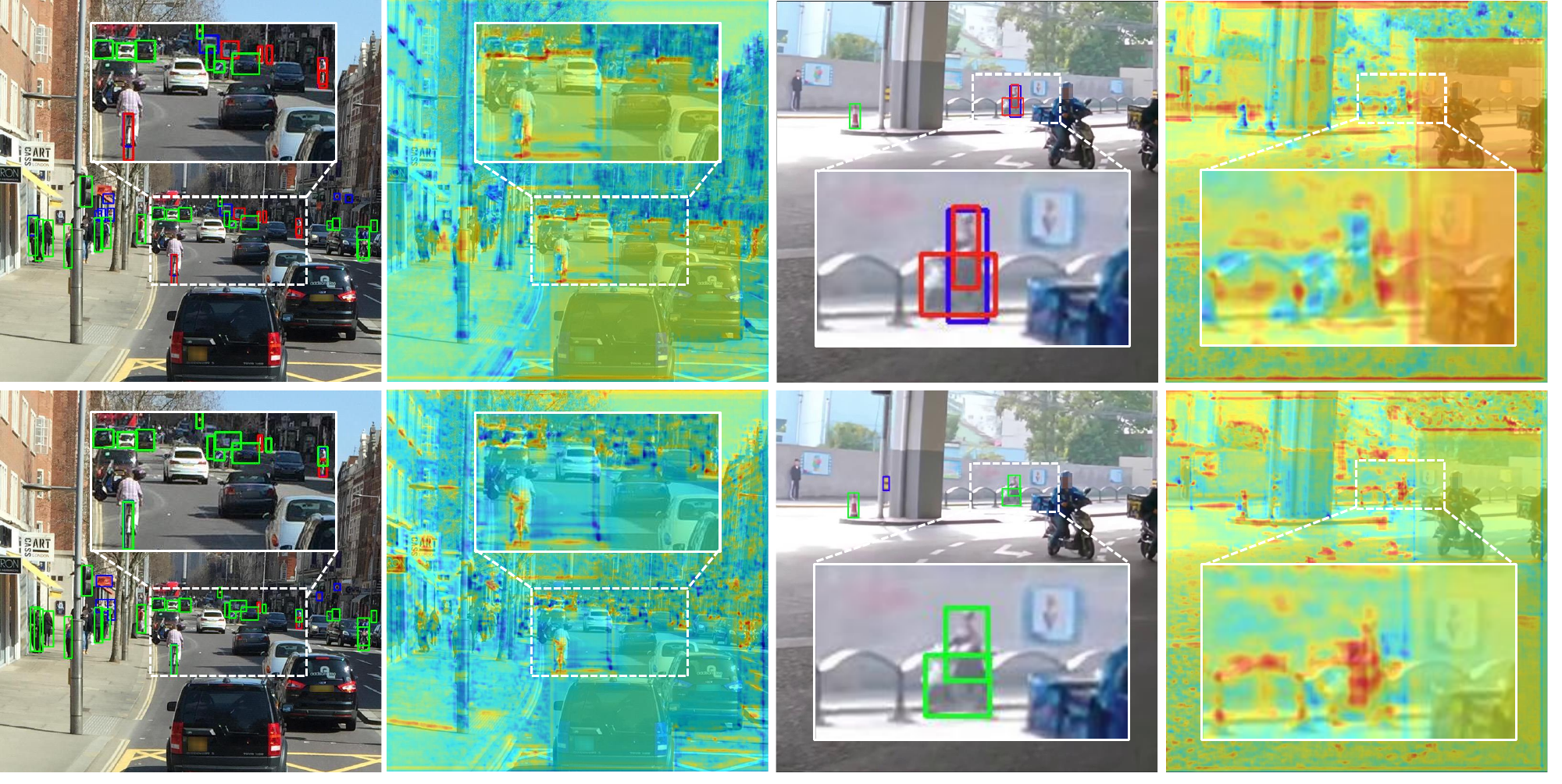}
\caption{Visualization of features map at the P2 level on SODA-D test set. The first row shows results from the baseline, while the second row presents results from DyFrDet. The green, blue, and red boxes denote true positives (TP), false positives (FP), and false negatives (FN), respectively. Please zoom in for details.}
\label{fig:heatmap}
% \vspace{-4mm}
\end{figure}

\noindent\textbf{Suppression Rate.}
We conduct an ablation study on different suppression rates, as shown in Table~\ref{tab:beta_ablation}. When $\beta = 0.00$, no suppression is applied. When $\beta = 1.00$, it corresponds to full filtering. The best performance is achieved when $\beta = 0.50$, indicating that moderate suppression yields an optimal trade-off between information retention and noise reduction. No suppression allows distractors to interfere with the target, while full suppression may mistakenly remove useful target features.

\subsection{Qualitative Visualization}
To qualitatively assess the effectiveness of our method, we visualize the feature heatmaps at the P2 level on SODA-D dataset, as shown in Fig.~\ref{fig:heatmap}. The heatmaps produced by CFINet exhibit cluttered and dispersed activations, which often lead to false positives and missed detections. In contrast, DyFrDet generates more concentrated activations on foreground regions, with suppressed responses in the background, resulting in more reliable detection performance, demonstrating its stronger ability to distinguish small objects from background clutter.

% \subsection{Complexity Comparison}
% Efficiency comparisons are provided in the supplementary material. Our method reduces GFLOPs compared with HS-FPN with only a 0.06$\times$ increase in parameters, and reduces both GFLOPs and parameters compared with DNTR~\cite{10518058}.

\section{Conclusion}
In this paper, we propose \textbf{DyFrDet}, a two-stage detector composed of a Dynamic Frequency-aware Feature Pyramid Network (DyFrFPN) and a Label Disambiguation Module (LDM). The former leverages Dynamic Band Predictor (DBP) to adaptively control the frequency representation, allowing the detector to suppress both low-frequency redundancies and high-frequency noises. The latter achieves accurate bounding box estimation by paying more attention to the high confidence samples and restraining the ambiguous samples. Experimental results demonstrate that our method achieves state-of-the-art performance on several widely used small object detection benchmarks.

\section{Acknowledgement} This work was supported by Project No. 25KYHX131B: Construction of Physical Neural Networks for Cognitive Countermeasures.
\bibliographystyle{ACM-Reference-Format}
% \bibliography{sample-base}
%%% -*-BibTeX-*-
%%% Do NOT edit. File created by BibTeX with style
%%% ACM-Reference-Format-Journals [18-Jan-2012].

\end{document}

%%
%% The "title" command has an optional parameter,
%% allowing the author to define a "short title" to be used in page headers.
\title{Supplementary Materials}

\renewcommand{\shortauthors}{Trovato et al.}

\maketitle

\section{Efficiency Comparison}
\begin{figure}[tbp]
\centering
\includegraphics[width=\linewidth]{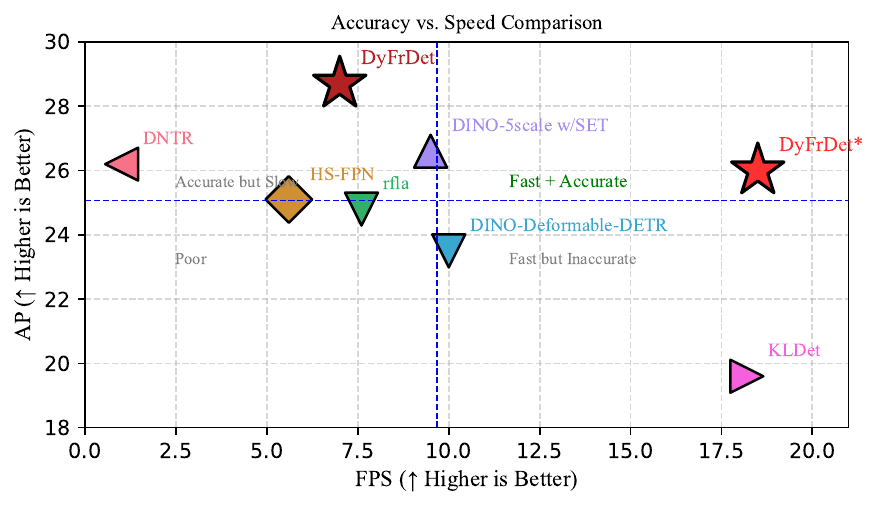}
\caption{Comparison with state-of-the-art methods on AI-TOD. We visualize Average Precision (AP) versus detection speed (FPS). Best viewed in color.}
\label{fig:speed_ap_comp}
\end{figure}

We evaluate inference speed on a single RTX3090. As illustrated in Fig.~\ref{fig:speed_ap_comp}, DyFrDet attains a runtime of 7.0 FPS, whereas DyFrDet$^{*}$ reaches 18.5 FPS. Under similar detection performance, DyFrDet$^{*}$ operates approximately 3.3$\times$ faster than HS-FPN (18.5 vs. 5.6 FPS) and 18.5$\times$ faster than DNTR (18.5 vs. 1.0 FPS). Even the standard DyFrDet achieves a speedup of 1.25$\times$ over HS-FPN (7.0 vs. 5.6 FPS), demonstrating the efficiency of our approach.

\section{Detection Performance Visualization}

\begin{figure*}[tbp]
\centering
\includegraphics[width=\linewidth]{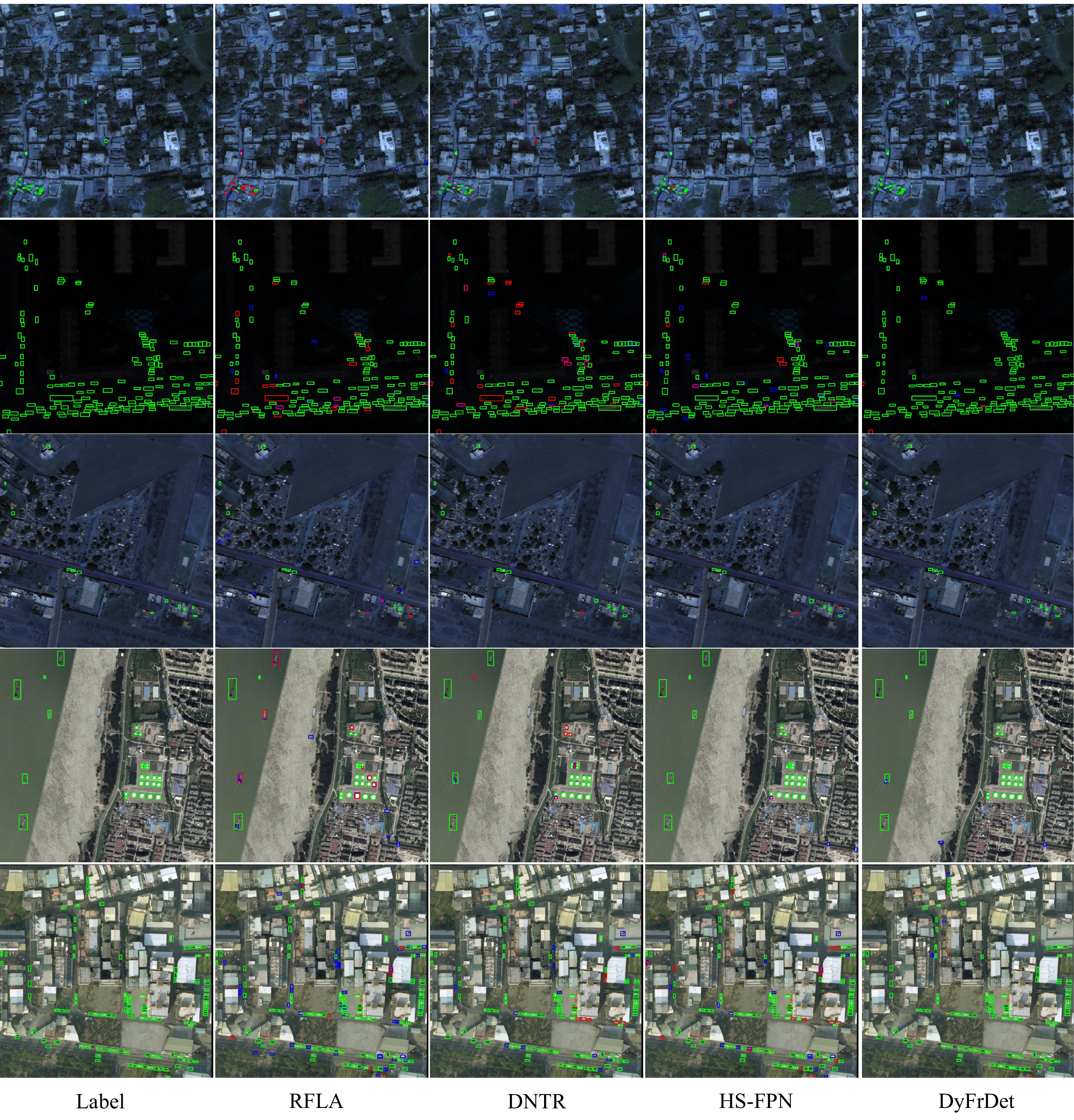}
\caption{Visualization of detection results of different object detection methods on the AI-TOD dataset. Green, blue, and red boxes denote true positives (TP), false positives (FP), and false negatives (FN), respectively.}
\label{fig:appendix_visualization_detection}
\end{figure*}

Fig.~\ref{fig:appendix_visualization_detection} presents qualitative comparisons between DyFrDet and three representative detection methods, with the IoU threshold uniformly set to 0.5. For clearer visualization, correctly detected objects (true positives) are marked with green bounding boxes, false detections (false positives) are indicated with blue boxes, and missed targets (false negatives) are highlighted in red.
The selected samples are drawn from the AI-TOD dataset and cover a variety of scenarios, including sparse scenes, dense scenes, and extremely low-light conditions. As shown in Fig.~\ref{fig:appendix_visualization_detection}, our method demonstrates superior detection performance across all scenarios, yielding more correct detections while significantly reducing false detections and missed targets.

\section{Heatmap Visualization}
\begin{figure*}[tbp]
\centering
\includegraphics[width=\linewidth]{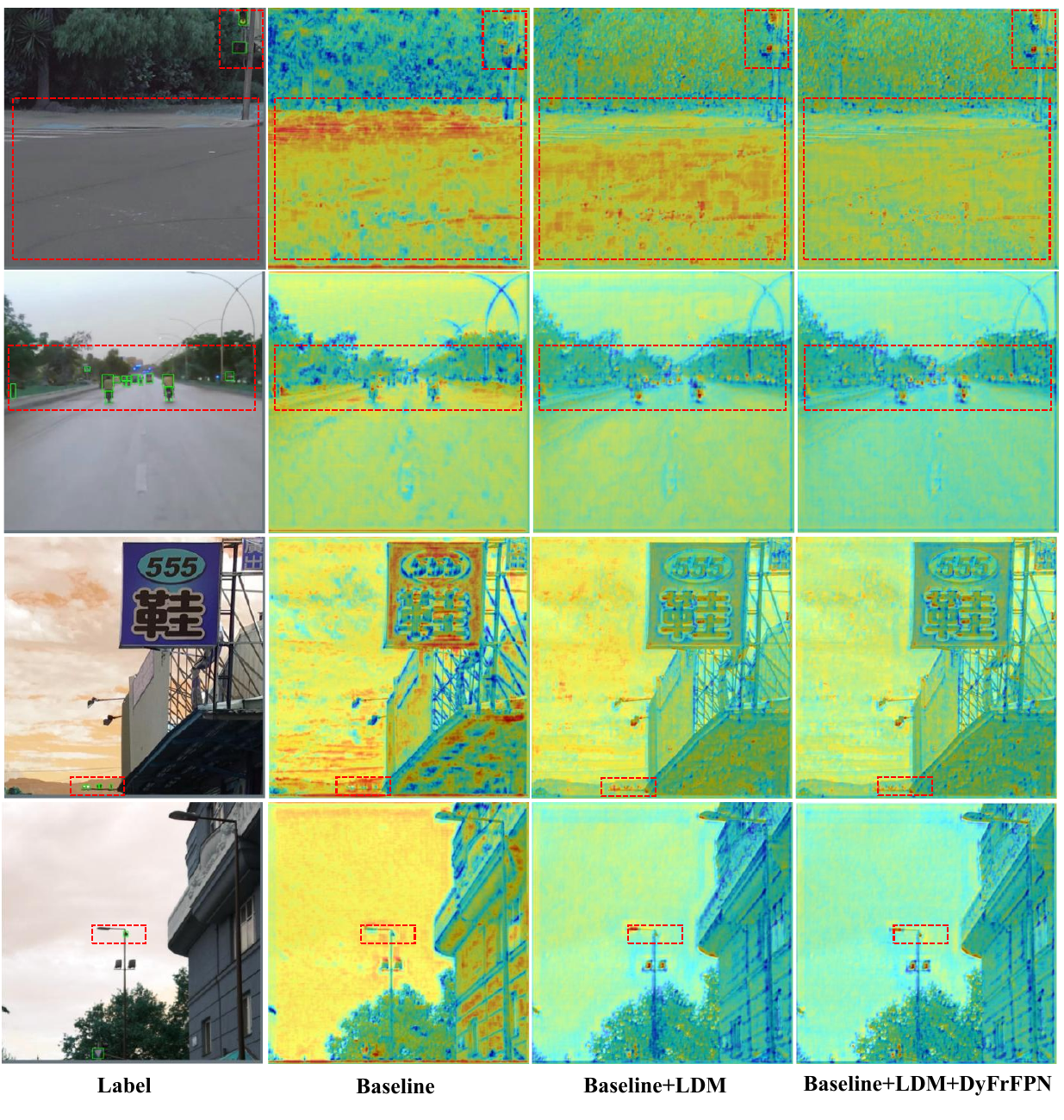}
\caption{Heatmap visualization on the SODA-D dataset. Key regions are highlighted in red. Please zoom in for details.}
\label{fig:ADD_MODULE}
\end{figure*}

To qualitatively validate the effectiveness of our design, we perform feature visualization by progressively incorporating the LDM and DyFrFPN modules. Specifically, we visualize feature maps at the P2 level of FPN using samples from the test split of SODA-D\cite{cheng2023towards} dataset.
As shown in Fig.~\ref{fig:ADD_MODULE}, the first column presents the ground-truth labels, the second column shows the baseline features, the third column corresponds to features after incorporating the LDM module, and the fourth column shows the features after integrating both LDM and DyFrFPN modules (i.e., DyFrDet). It can be observed that the features become increasingly discriminative with the progressive introduction of our designed module. Concurrently, the background responses are gradually suppressed, while foreground regions remain highly activated, which validates the effectiveness of our proposed methods.

\section{Heatmap Comparison}
\label{appendix:heatmap_visualization}
\begin{figure*}[tbp]
\centering
\includegraphics[width=\linewidth]{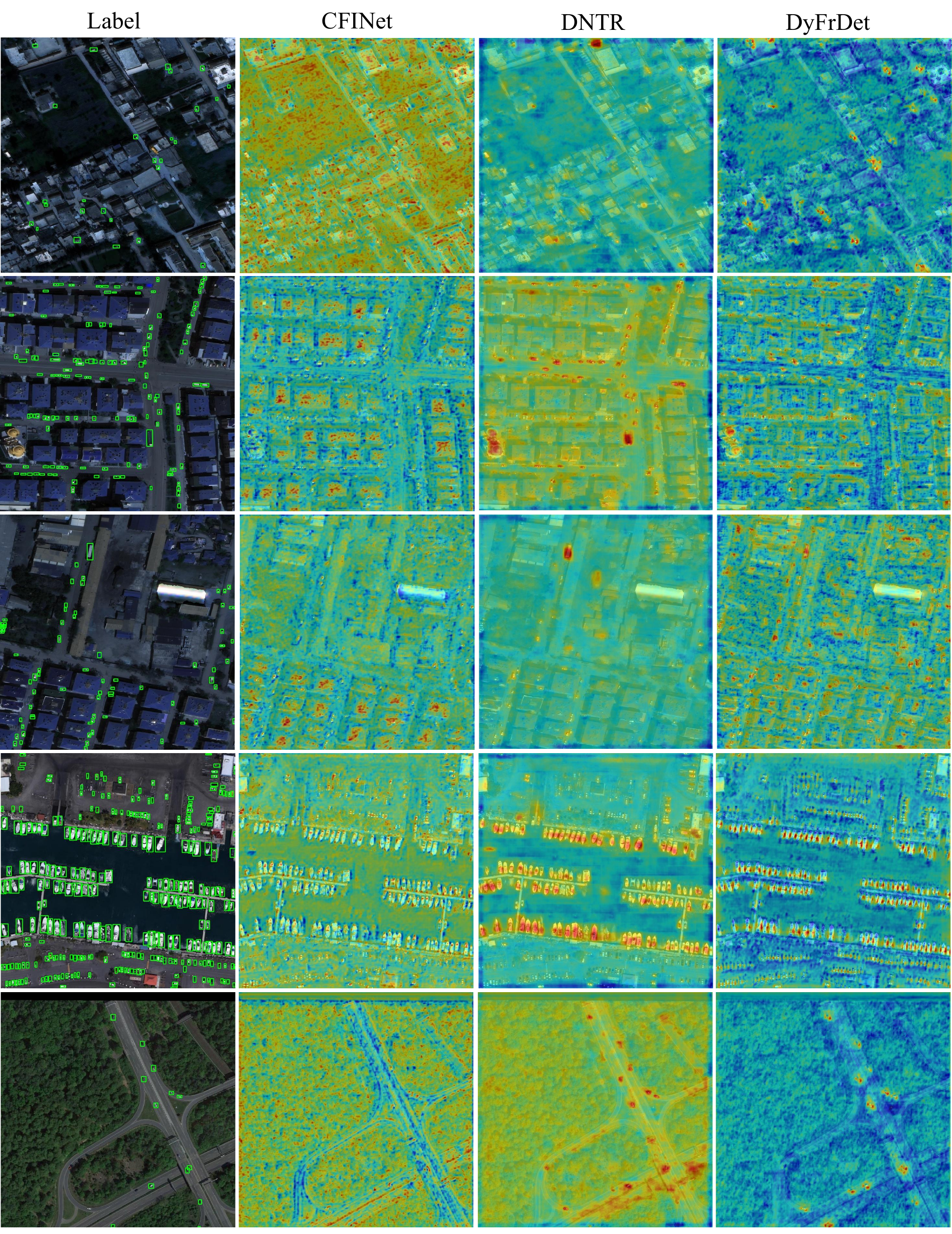}
\caption{Comparison of feature maps at the P2 level on the AI-TOD test set against other SOTA methods.}
\label{fig:heatmap_comparison}
\end{figure*}

To qualitatively validate the effectiveness of the proposed method, we present additional heatmap comparisons with several state-of-the-art (SOTA) approaches on AI-TOD.
As shown in Fig.~\ref{fig:heatmap_comparison}, we compare our method with CFINet~\cite{CFINet} and DNTR~\cite{10518058}. We visualize feature maps at the P2 level of FPN. The first column shows the ground-truth labels, while the second, third, and fourth columns correspond to the features produced by CFINet, DNTR, and our method, respectively. Compared with these methods, DyFrDet generates more fine-grained and discriminative feature representations, which demonstrates its effectiveness.

\section{More Visualization with Different $\sigma_m$}
\begin{figure*}[tbp]
\centering
\includegraphics[width=\linewidth]{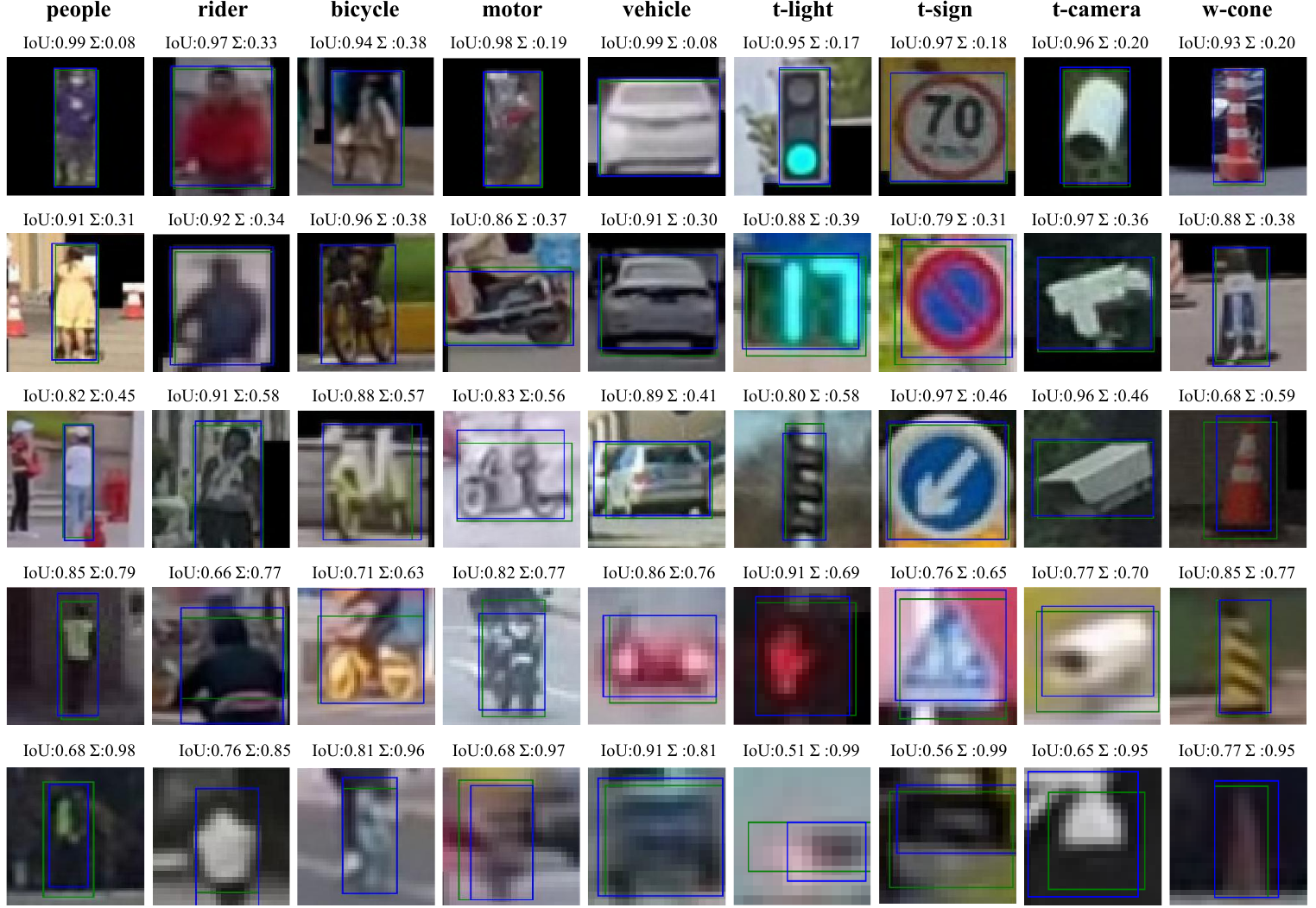}
\caption{Visualization of different instances with varying $\sigma_m$ across classes on the SODA-D dataset. \textit{t-light, t-sign, t-camera, w-cone} are abbreviations for traffic light, traffic sign, traffic camera, and warning cone, respectively.}
\label{fig:appendix_visualization_sigma}
\end{figure*}

Here, we provide additional visualizations to illustrate the effect of $\sigma_m$. We select samples where the predicted boxes have an IoU greater than 0.5 with the ground-truth labels. As shown in Fig.~\ref{fig:appendix_visualization_sigma}, green boxes denote ground-truth labels, while blue boxes indicate predictions. 
It can be observed that as $\sigma_m$ increases, instances of different categories gradually become more blurred. When $\sigma_m$ approaches 1, some samples exhibit mislabeling, such as the traffic light, traffic sign, and traffic camera in the last row. These visualizations qualitatively demonstrate the reasonableness of using $\sigma_m$ to guide sample classification.

\section{Computing Infrastructure}
The experiments were conducted in the following environment to ensure reproducibility:

\begin{itemize}
  \item \textbf{GPU:} NVIDIA RTX 3090
  \item \textbf{OS Version:} Ubuntu 22.04 LTS
  \item \textbf{CUDA Version:} 11.8
  \item \textbf{Python Version:} 3.8
  \item \textbf{PyTorch:} 1.11.0+cu113
  \item \textbf{Torchvision:} 0.12.0+cu113
\end{itemize}

\bibliographystyle{ACM-Reference-Format}
%%% -*-BibTeX-*-
%%% Do NOT edit. File created by BibTeX with style
%%% ACM-Reference-Format-Journals [18-Jan-2012].